# FaceDig: Automated tool for placing landmarks on facial portraits for geometric morphometrics users

Karel Kleisner[1]*, Jaroslav Trnka[1], Petr Tureček[1]

[1] Department of Philosophy and History of Science, Faculty of Science, Charles University, Prague, Czech Republic

* Corresponding author: karel.kleisner@natur.cuni.cz

Abstract

Landmark digitization is essential in geometric morphometrics, enabling the quantification of biological shapes, such as facial structures, for in-depth morphological analysis. Traditional landmarking, which identifies specific anatomical points, can be complemented by semilandmarks when precise locations are challenging to define. However, manual placement of numerous landmarks is time-consuming and prone to human error, leading to inconsistencies across studies. To address this, we introduce FaceDig, an AI-powered tool designed to automate landmark placement with human-level precision, focusing on anatomically sound facial points. FaceDig is open-source and integrates seamlessly with analytical platforms like R and Python. It was trained using one of the largest and most ethnically diverse face dataset, applying a landmark configuration optimized for 2D enface photographs. Our results demonstrate that FaceDig provides reliable landmark coordinates, comparable to those placed manually by experts. The tool's output is compatible with the widely-used TpsDig2 software, facilitating adoption and ensuring consistency across studies. Users are advised to work with standardized facial images and visually inspect the results for potential corrections. Despite the growing preference for 3D morphometrics, 2D facial photographs remain valuable due to their cultural and practical significance. Future enhancements to FaceDig will include support for profile views, further expanding its utility. By offering a standardized approach to landmark placement, FaceDig promotes reproducibility in facial morphology research and provides a robust alternative to existing 2D tools.

Keywords

Landmark Digitization; Geometric Morphometrics; Facial Morphology; Automated Landmarking; Morphometric Software; 2D Facial Analysis

## 1. Introduction

Landmark digitization is a fundamental step in geometric morphometrics, allowing researchers to quantify and analyze the geometric aspects of biological structures, such as bones, organisms, or other objects. Landmark-based geometric morphometrics captures morphological information, enabling the assessment of covariation between shape and other external or internal factors.

Landmark digitization typically involves the precise identification of specific anatomical points on an object of interest. These landmarks serve as reference points that characterize the object's shape. When true landmarks cannot be identified, semilandmarks (or sliders) can be used instead. Semilandmarks usually refer to points that are not precisely located at anatomically or geometrically well-defined locations on the object. Unlike traditional landmarks that have clear, homologous positions on each specimen, semilandmarks are positioned along curves or outlines and capture shape information in regions where defining exact landmarks may be challenging. Semilandmarks provide additional shape information, especially in areas where traditional landmarks may not be practical or informative. This is particularly useful when studying complex structures or soft tissues. However, semilandmarks are more prone to variation due to digitizing error and should therefore not be treated in the same way as true landmarks.

Manually applying numerous landmarks and semilandmarks to a large number of facial photographs is a time-consuming process that many researchers hesitate to undertake. Moreover, scientists vary in their ability to detect homologies and in the precision of placing landmarks in corresponding locations across objects in the series.

Some recent works use landmark configurations originally designed for stimuli manipulation in software such as PsychoMorph, rather than for biometric purposes, e.g., (Antar & Stephen, 2021; Marcinkowska et al., 2021; Marcinkowska & Holzleitner, 2020). Landmark definitions and placement should ideally adhere to criteria of anatomical correspondence, such as the classification based on Bookstein's typology of landmarks (Bookstein, 1991). In practice, such criteria are often relaxed (Wärmländer et al., 2019), but one should definitely maintain a high level of landmark precision to ensure a reasonable degree of anatomical correspondence across all objects in the set.

We have decided to relieve face researchers of these manual tasks by training an AI agent named FaceDig, which achieves human-level precision while focusing on anatomically sound facial points. FaceDig is open-source and freely available at facedig.org. It can be easily loaded and used in any analytical software, including R and Python. When provided with standardized facial photographs, FaceDig returns landmark coordinates in a format identical to that of the popular landmarking software TpsDig2 by James Rohlf (Rohlf, 2015, 2021).

FaceDig was trained to apply the landmark configuration used in our previous research (citations of our papers) and is similar to approaches reflected in the work of other geometric morphometrics practitioners (e.g., Fink et al., 2005; Mitteroecker et al., 2013, 2015; Schaefer et al., 2005, 2006). We believe that our configuration effectively captures the morphological information available in 2D enface photographs. Our training dataset of faces was manually landmarked with high precision and is one of the largest and most ethnically diverse in the current literature.

## 2. Methods

### 2.1. Landmark description

Applying geometric morphometrics to human facial shape involves the precise use of landmarks to capture facial morphology. Ideally, landmarks should denote key morphological features as discrete, anatomically homologous points that are identifiable and reproducible across all specimens (James Rohlf & Marcus, 1993). Landmarks can also be located at points of maximum or minimum curvature or other geometric properties of the object, or defined by their relative position to other landmarks (Bookstein, 1991). They may lie on the boundaries or edges of anatomical structures. The description of shape can be further supplemented by the use of semilandmarks (Bookstein, 1997; Gunz & Mitteroecker, 2013). These are placed to capture the shape of outlines and curved surfaces when the continuous nature of a morphological structure resists delimitation into discrete structures or boundaries describable by fixed anatomical points. FaceDig places both landmarks and semilandmarks on the human face but does not qualitatively differentiate between them. Points that should be treated as Semilandmarks therefore need to be defined a posteriori (our examples contain *slidsR* file that allows *geomorph* R package or other similar software to do this), after landmark delineation and before entering the Generalized Procrustes Analysis (GPA). Table 1 provides the definitions of both landmarks and semilandmarks. Their placement on the face is depicted in Figure 1.

Table 1. Definitions of landmarks and semilandmarks on the human face

| No | Name | Description |
|---|---|---|
| 1 | TRICHION | midpoint of the hairline, that is, on the hairline through the midline of the forehead |
| 2 | MENTON | the lowest point of the lower border of the mandible (along the jaw line) |
| 3 | LABIALE INFERIUS | the midline point of the lower vermilion line (border of the lower lip) |
| 4,5 |  | the midpoints between LABIALE INFERIUS (3) and CHEILON (9,10) |
| 6,7 | CHEILON | the outer corner of the mouth where the outer edges of the upper and lower lip meet |
| 8 | LABIALE SUPERIUS | the upper midpoint of the upper vermilion line, a point of maximum local curvature between the christae philtri |
| 9,10 | CHRISTA PHILTRI | the point on the crest of the philtrum, the vertical groove in the median portion of the upper lip on the vermilion border |
| 11,12 |  | the midpoints between LABIALE SUPERIUS (8) and CHRISTA PHILTRI (9,10) |
| 13 | SUBNASALE | midpoint of the angle at the columella base where the lower border of the nasal septum and the surface of the upper lip meet |
| 14,15 | COLUMELLA APEX | highest point of the columella crest at the apex of the nostril |
| 16,17 | ALARE | the most lateral point on the ala contour |
| 18,19 | ALAE ORIGIN | the most posterolateral point of the curvature of the base of the nasal alae |
| 20,27 | ENDOCANTHION | the inner corner of the eye fissure where eyelids meet |
| 21,28 | EXOCANTHION | the outer corner of the eye fissure where eyelids meet |
| 22,30 | PALPEBRALE SUPERIUS | the highest visible point of the iris |
| 23,29 | PALPEBRALE INFERIUS | the lowest visible point of the iris |
| 24,31 | Iris Outer Border | the rightmost point of the right iris (leftmost of the left iris) |
| 25,32 | Iris Inner Border | the leftmost point of the right iris (rightmost of the left iris) |
| 26,33 | | the midpoint between ENDOCANTHION (20,27) and PALPEBRALE INFERIUS (23,29) |
| 34,36 | SUPERCILIARE LATERALE | the most lateral point of the eyebrow |
| 35,37 | SUPERCILIARE MEDIALE | the most medial point of the eyebrow |
| 38-40,44-46 | the eyebrow upper curve | three semilandmarks with regular spacing between SUPERCILIARE LATERALE (34,36) and SUPERCILIARE MEDIALE (35,37) |
| 41-43,47-49 | the eyebrow lower curve | three semilandmarks with regular spacing between SUPERCILIARE LATERALE (34,36) and SUPERCILIARE MEDIALE (35,37) |
| 50,59 | ZYGION | the most lateral point of the zygomatic arch |
| 51-58,60-67 | the lower jaw | eight semilandmarks with regular spacing between MENTON (2) and ZYGION (50,59) |
| 68 | STOMION | center of the lip crack, lying on the midline between LABIALE SUPERIUS (8) and LABIALE INFERIUS (3) |
| 69,70 | | the midpoints between STOMION (68) and CHEILON (6,7) |
| 71,72 | Pupil | center of the pupil |

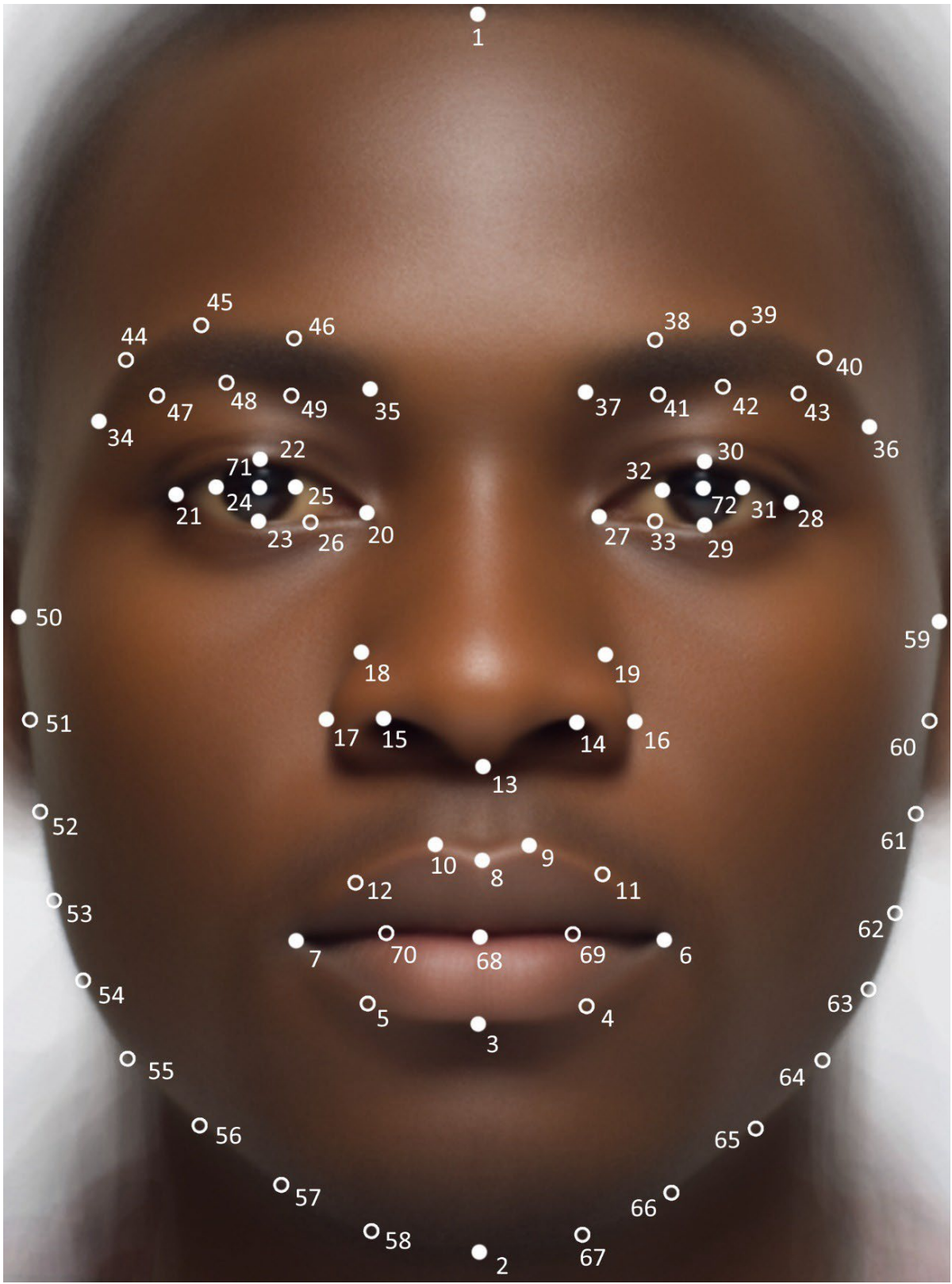

Figure 1. Illustration of the landmark and semilandmark positions on the human face. Landmarks are marked as filled circles, while semilandmarks are marked by empty circles.

2. 2. Data input and digitizing

A total of 3,937 photographs (1,955 women, 1,982 men) from 14 distinct cultures were used in the training. A detailed breakdown of the sample by population/dataset and sex can be found in Table 2.

Table 2. Facial databases used in the training of FaceDig mostly contain facial photographs that were manually landmarked for the purposes of our previous research. References are provided where available.

| DATASET | Women | Men | Total | Reference |
|---|---|---|---|---|
| Cameroon | 201 | 188 | 389 | (Kleisner et al., 2017, 2021, 2024) |
| Colombia | 220 | 202 | 422 | (Kleisner et al., 2021, 2024) |
| Czech | 818 | 704 | 1522 | (Kleisner et al., 2010, 2021, 2024) |
| India-CFD | 52 | 90 | 142 | (Lakshmi et al., 2021) |
| India-Nalagarh | 62 | 60 | 122 | unpublished |
| Mauritania | 83 | 58 | 141 | unpublished |
| UFC faces | 0 | 146 | 146 | (Třebický et al., 2013) |
| Namibia | 53 | 49 | 102 | (Kleisner et al., 2017, 2021, 2024) |
| North Africa | 0 | 50 | 50 | (Courset et al., 2018) |
| Poland | 152 | 55 | 207 | (Danel et al., 2016; Marcinkowska et al., 2020) |
| Romania | 50 | 50 | 100 | (Kleisner et al., 2021, 2024) |
| Senegal | 49 | 50 | 99 | unpublished |
| Türkiye | 115 | 144 | 259 | (Saribay et al., 2021) |
| United Kingdom | 50 | 50 | 100 | (Kleisner et al., 2021, 2024) |
| Vietnam | 50 | 86 | 136 | (Kleisner et al., 2024; Pavlovič et al., 2023) |
| **SUM** | **1955** | **1982** | **3937** | |

2.3. Model Structure and Training

Our model integrates several pre-existing components and custom deep learning architecture to enhance the precision of facial landmark detection. Initially, we employed the MediaPipe library (Lugaresi et al., 2019) for its face detection and landmark detection capabilities. Although MediaPipe provides a robust initial framework, its precision is insufficient for our purposes. Therefore, it serves as an auxiliary input, which is further refined by our model.

The workflow of our model is as follows:

1. Initial Landmark Projection: The facial landmarks detected by MediaPipe are fed into the first component of our deep learning model. This component functions as an ensemble projection layer, transforming the MediaPipe landmark system into our landmark configuration.
2. Convolutional Neural Network (CNN) Enhancement: The transformed landmarks from the projection layer are then processed by the CNN (Lecun et al., 1998) component of our model. This CNN processes the image or specific cropped regions of the image, enhancing the landmark precision. The CNN focuses on refining these rough projections to pinpoint accurate facial landmarks.

The complete model architecture, which combines the pre-trained MediaPipe package, a linear projection layer, and a CNN Enhancement, was trained using a substantial dataset of over 5,000 landmark configurations (based on facial photographs of 3,937 individuals; some of faces were landmarked twice). This extensive training set ensures the model's robustness across diverse facial features and conditions.

We adopted a sequential training approach for model optimization:

1. Projection Layer Training: Initially, the projection layer was trained for 150 epochs to establish a preliminary transformation of the MediaPipe landmarks into our system.
2. CNN Training: Subsequently, the CNN was trained for an additional 20 epochs, refining the preliminary landmarks to achieve high precision.

This sequential training strategy allowed for the effective integration and enhancement of the MediaPipe's initial detections, resulting in a highly accurate facial landmark detection model.

2.4. Short user manual

The model is available for free download as a relatively small portable application that does not require any additional setup. Since the model still exceeds 1 GB in size, users should ensure they have sufficient bandwidth and storage capacity before proceeding with the download.

To use the application, users simply need to browse their folder system and select the folder containing the images they wish to process. The model then processes all images in the selected folder, applies precise facial landmark detection, and returns a single .tps dataset similar to the output of TpsDig2 software for manual landmark digitization. For a detailed version of the manual with images, visit facedig.org

2. 5. Repeatability

In the case of manually applied landmarks, digitizing should be performed multiple times by the same researcher (or by different individuals) to assess the accuracy of landmark positioning, referred to as inter-digitizer repeatability. Consistency in landmark identification is crucial for reliable results.

To assess whether automatic landmarks closely approximate manually placed landmarks, we calculated inter-digitizer repeatability using the same set of facial portraits. We employed a set of 100 artificial faces generated by NVIDIA's generative adversarial network, StyleGAN2, from our previous study (Boudníková & Kleisner, 2024). We first generated two files of coordinates landmarked by FaceDig. The landmark configurations in the first file (hereafter referred to as FDG) were kept as they were output by FaceDig, without any human intervention. The second file was manually corrected by a human digitizer (hereafter referred to as HC) using the software TpsDig2 to achieve the same (or higher) precision in landmark configuration as in our previous studies where the landmarks were applied manually.

Moreover, we compared two basic morphometric measures—distinctiveness and asymmetry—derived from automatically and manually placed landmarks (FDG and HC, respectively). Asymmetry was calculated so that the aligned coordinates after Procrustes fit were first laterally reflected along the midline axis. The corresponding paired landmarks on the left and right sides of the faces were then relabeled, swapping the numeric labels of landmarks on the left side with those on the right side and vice versa. Procrustes distances between the original and the mirrored (reflected and relabeled) configurations were then calculated, with larger values indicating greater facial asymmetry. Distinctiveness was computed separately for FDG and HC coordinates aligned by Procrustes fit as the Procrustes distance from the group mean. The greater the distance, the more distinct (and less average) the face.

## 3. Results

Using *procD.lm ()* function, we performed a Procrustes analysis of variance between individual faces and two replicates (i.e., FDG and HC) of the same face. Repeatability was quantified as the ratio of two variance components: the interindividual variance and the sum of interindividual variance with variance of the individual:replicate interaction. The repeatability of digitizing precision between the two replicates was 0.97.

Based on the Procrustes residuals of facial configurations, we calculated distinctiveness separately for FDG and HC replicates as the Procrustes distance from the group mean. The Pearson correlation between the replicates was above 0.9. (n =100, r = 0.94, p < 0.001, 95% CI [0.92, 0.96]).To estimate the impact of repeatability on the assessment of facial symmetry, we calculated asymmetry scores separately for each of the FDG and HC replicates. The correlation between the asymmetry scores of the FDG and HC replicates was tight (n =100, r = 0.92, p < 0.001, 95% CI [0.88, 0.94]), indicating a high level of precision. Figure 2 contains graphical representation of this close relationship.

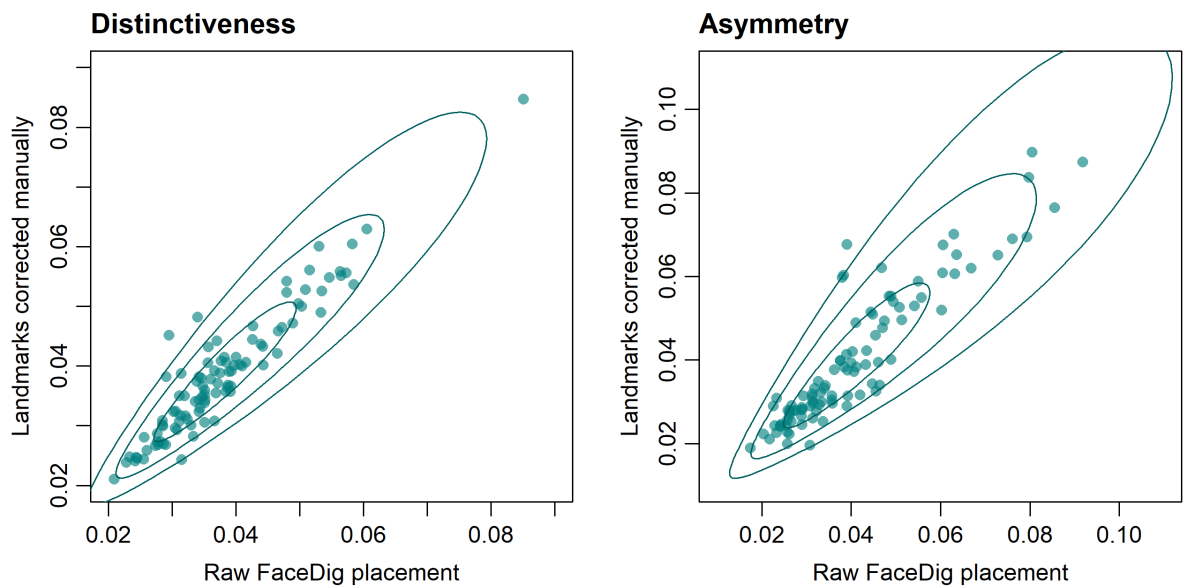

Figure 2. Composite measures based on raw FaceDig landmark placement, and it's manually refined counterpart. The contours outline 50%, 90% and 99% of expected points and are based on logarithm covariation. This prediction is technically more appropriate than one based on Pearson's correlation because both plotted distance measures have a meaningful 0 and cannot

assume negative values. Correlations of Distinctiveness and Asymmetry logarithms (0.93 [95% CI: 0.90, 0.95] and 0.91 [95% CI: 0.87, 0.94] respectively) are, however, almost identical to correlations calculated from the non-transformed data.

## 4. Conclusions

Over the last few decades, there has been a substantial rise in the adoption and advancement of landmark-based geometric morphometric techniques. Face research and the study of facial shape variation have extensively incorporated these methodologies. Here, we demonstrate that our open-source tool, FaceDig, achieves landmark placement precision comparable to that of human researchers. We strongly recommend that users provide FaceDig with standardized facial images to achieve optimal results, as the tool was trained on standardized enface photographs. Users should also visually inspect all outputted facial configurations and manually correct any apparent landmark displacements (using TpsDig2, for example) caused by software misinterpretation. Our results, however, demonstrate that even the raw FaceDig output allows for the extraction of reliable summary characteristics such as distinctiveness and asymmetry.

FaceDig is optimized for facial shape analysis, representing an alternative to existing software operating on 2D photographs (Jones et al., 2021). Despite the preference for 3D morphometrics in physical anthropology, photographs of human faces retain their significance in human culture. Photographs are omnipresent on the internet, social networks, and various medicinal and institutional databases, representing an immense amount of visual information that can potentially be analyzed.

In the near future, we plan to enhance FaceDig's functionality to include digitization of profile facial photographs. Combining information from both enface and profile views will further improve the overall performance of facial morphometrics. We also plan to release updated and improved versions of the app at regular intervals and hope that future versions of FaceDig will demonstrate even higher fidelity in landmark delimitation on human faces

FaceDig provides landmark configurations identical to (or very similar to) those used in many previous studies on facial morphology. Introducing this tool to the scientific community thus increases the repeatability of research in our field.

**Acknowledgements**

This study was supported by the Czech Science Foundation project, reg. no 24-11735S.

**Ethics statement**

The study was performed in accordance with the ethical standards as laid down in the Helsinki Declaration. All procedures mentioned and followed were approved by the Institutional Review Board of the Faculty of Science of the Charles University (protocol ref. number 06/2023).

**Data and code availability**

All data and code that allow the reproduction of the analysis and visualization are available at osf.io/wc2em

The program is hosted at facedig.org.